\documentclass{article}

    \PassOptionsToPackage{numbers, compress}{natbib}
\usepackage[preprint]{neurips_2026}

\usepackage[utf8]{inputenc} % allow utf-8 input
\usepackage[T1]{fontenc}    % use 8-bit T1 fonts
\usepackage[colorlinks,linkcolor=black,anchorcolor=black,citecolor=black]{hyperref}       % hyperlinks
\usepackage{url}            % simple URL typesetting
\usepackage{booktabs}       % professional-quality tables
\usepackage{amsfonts}       % blackboard math symbols
\usepackage{nicefrac}       % compact symbols for 1/2, etc.
\usepackage{microtype}      % microtypography
\usepackage{xcolor}         % colors
\usepackage{wrapfig}

\usepackage{amsmath}
\usepackage{amssymb}
\usepackage{mathtools}
\usepackage{amsthm}
\usepackage{multirow}
\usepackage[capitalize,noabbrev]{cleveref}

\usepackage{color}

\usepackage{xspace}
\newcommand{\name}[0]{MBP-KT\xspace}

\makeatletter
\DeclareRobustCommand\onedot{\futurelet\@let@token\@onedot}
\def\@onedot{\ifx\@let@token.\else.\null\fi\xspace}

\def\eg{\emph{e.g}\onedot}

\makeatother

\usepackage{algpseudocode}  
\usepackage{algorithm}
\theoremstyle{plain}

\theoremstyle{definition}

\theoremstyle{remark}

\usepackage[textsize=tiny]{todonotes}

\title{\name: Learning Global Collaborative Information from Meta-Behavioral Pattern for Enhanced Knowledge Tracing}

\author{
  Yuhao Jia$^{1}$\thanks{Equal contribution.}, 
  Duantengchuan Li$^{2}$\footnotemark[1],
  Jinsong Chen$^{1}\thanks{Corresponding author.}$,\\
  \textbf{Zhongjie Mao}$^{3}$\textbf{,} 
  \textbf{Mingwen Tong}$^{1}$\footnotemark[2]\textbf{,}  
  \textbf{Yue Li}$^{2}$\textbf{,}  
  \textbf{Xiaoguang Wang}$^{2}$\\
  $^1$Faculty of Artificial Intelligence in Education, Central China Normal University\\
  $^2$School of Information Management, Wuhan University\\
  $^3$School of Big Data and Artificial Intelligence, Chizhou University\\
  \texttt{yhj1126@mails.ccnu.edu.cn, guangnianchenai@ccnu.edu.cn}
}

\begin{document}

\maketitle

\begin{abstract}
The emerging collaborative information-based knowledge tracing (KT) has been a promising way to enhance modeling of learners' knowledge states.
The core idea is to extract the collaborative information from interaction sequences of other learners to assist the prediction on the target one.
Despite effectiveness, existing methods are built on the raw interaction sequences with tailored modules, which inevitably limits their capacity in deeply capturing learning behavioral patterns and generalization.
To this end, we propose a general meta-behavioral pattern-aware framework (MBP-KT) for KT.
Specifically, MBP-KT introduces a novel meta-behavioral sequence construction to transform the raw interaction sequences into the combinations of different meta-behavioral patterns.
In this way, the learning behavioral patterns of learners can be effectively preserved.
Then, MBP-KT develops a parameter-free module to extract the global collaborative representations from the constructed meta-behavioral sequences.
Moreover, MBP-KT provides general injection strategies to introduce the extracted global collaborative information into various downstream KT models, ensuring the universality of the collaborative information.
Extensive results on real-world datasets demonstrate that MBP-KT can consistently boosts the performance of a wide range of KT models.
\end{abstract}

%One sentance
%we propose a new framework MBP-KT that extracts the global collaborative information from the constructed meta-behavior sequences to enhance the performance of various KT models.

%key words
%knowledge tracing, collaborative information, meta-behaviroal pattern, meta-behavioral sequence

\section{Introduction}
\label{Sec:intro}

Knowledge Tracing (KT) is a fundamental task in educational data mining~\cite{abdelrahman2023knowledge, manouselis2012recommender}, aiming to predict the performance of the target learner on the next exercise according to the historical response logs.
Since the response logs of learning interactions in knowledge tracing scenarios naturally exhibit sequential characteristics, various sequential models have been widely applied to this task over the past decade~\cite{romero2020educational}.
The core idea of these methods is to utilize the advanced sequential models, such as recurrent neural networks (RNNs)~\cite{piech2015deep, yin2019quesnet}, Transformer~\cite{ghosh2020context, shin2021saint, pandey2020rkt} and state space models (SSMs)~\cite{liu2024mamba4kt}, to learning the evolving knowledge states from the individual's learning sequence, achieving remarkable results.

Nevertheless, modeling based on individual sequence suffers from inherent limitations.
Deeply mining an individual learner's interaction records inevitably overlooks the common learning patterns that emerge from other learners' interaction records.
In educational contexts, these common learning patterns play a non-negligible role~\cite{long2022improving, wang2019neural, he2020lightgcn} in modeling learners' knowledge states.
Hence, recent studies~\cite{minn2018deep,long2022improving,CoSKT, yang2021enhancing} attempt to extract the collaborative information from interactions of other learners for improving the performance for KT.

The core difference among these methods lies in how they extract collaborative information from the interaction records of learners. 
\cite{minn2018deep} computes learners' response performance on each knowledge concept, and further employs clustering strategies~\cite{lin2021improving} to identify groups of learners with similar performance. \cite{long2022improving} and \cite{CoSKT}, on the other hand, leverage subsequence matching based on the target learner's historical response sequence to mine groups of learners who exhibit similar response sequences.

Despite effectiveness, existing methods are built on the raw interaction sequences that identifies behaviorally similar learner based on the specific exercises or knowledge concepts, making collaborative signals restricted by the explicit content overlap.
On the one hand, such a content-bound matching strategy is easily affected by the unstable data issues of the raw interaction sequences (\eg, data sparsity)~\cite{lee2022contrastive, choffin2019das3h} which are frequently encountered in real-world applications.
On the other hand, these strategies fail to capture the deeper behavioral commonalities that transcend specific content, thereby restricting the extraction of meaningful collaborative information~\cite{gao2024collaborative, zhang2022multi}.
Besides the limitations in extraction of collaborative information, existing paradigms also lack architectural generality. As a global context feature, collaborative information extracted from all interaction responses should have the potential to serve various downstream models~\cite{vie2019knowledge, ghosh2021bobcat}. However, current methods typically tailor their collaborative modules to specific neural network backbones, which restricts the global collaborative information from being universally applied to other diverse downstream models.
Therefore, a research question naturally arises:
\textit{How can we effectively extract the collaborative information beyond raw interaction sequences and utilize this to serve various downstream KT methods?}

\begin{figure}[t]
    \centering
    \includegraphics[width=13cm]{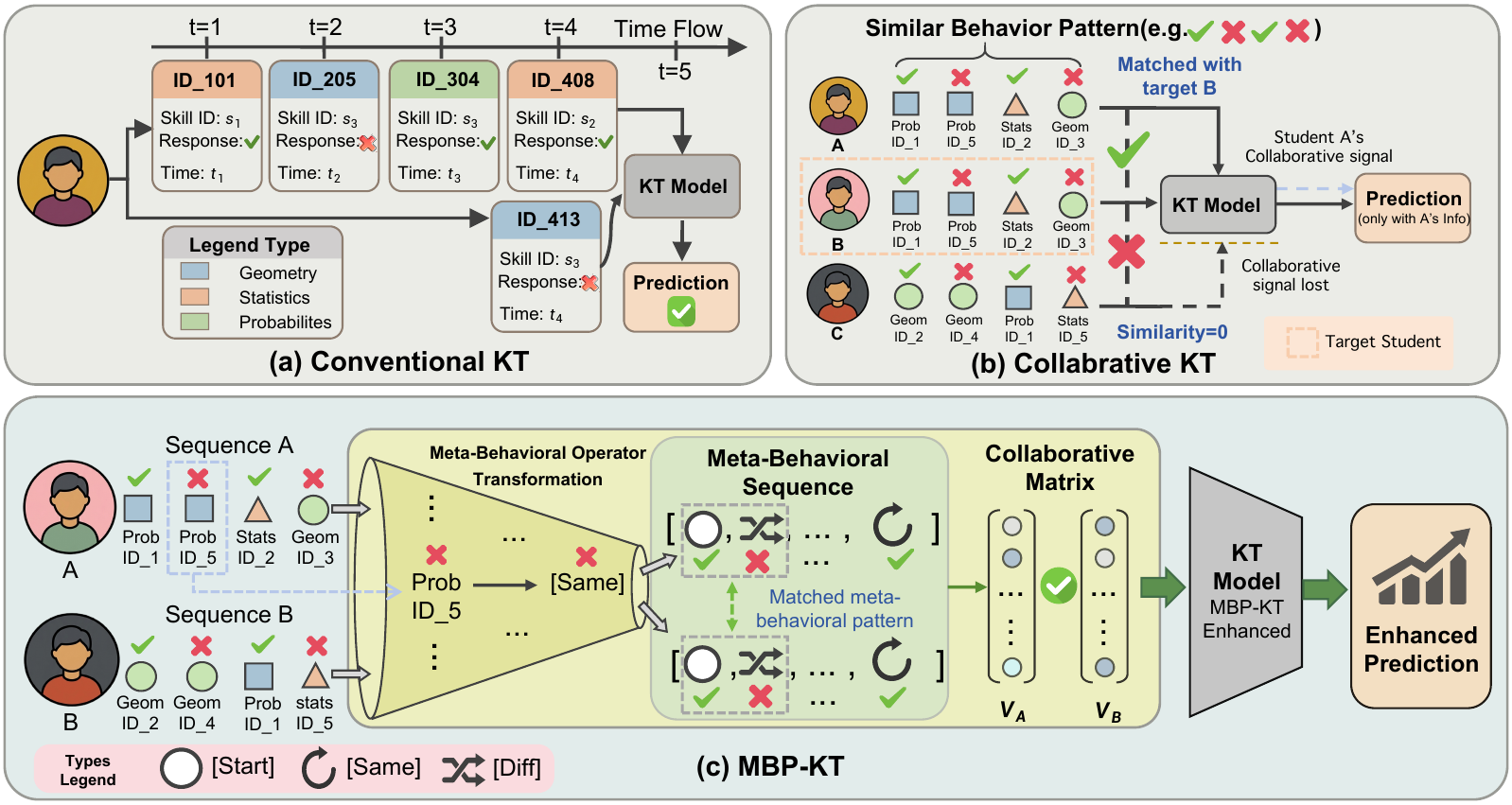}
    \caption{
    The comparison between \name and previous methods. 
    (a) Conventional KT models only leverage the individual responses for prediction. 
    (b) Prior collaborative KT models rely on the behavior match through the raw interaction sequences to extract the collaborative information for serving the specific model. 
    (c) \name transforms raw sequences into meta-behavioral sequences to extract the global collaborative features for enhancing the performance of diverse KT models.}
    \label{fig:intro}
\end{figure}

To answer this question, we propose a general meta-behavioral pattern-aware framework for KT (termed \name).
\autoref{fig:intro} illustrates a comparison between \name and prior methods.
Different from previous methods, \name develop a novel meta-behavioral sequence construction strategy that leverages the delicately designed behavior operators to transform the raw interaction sequences into meta-behavioral sequences.
This strategy decouples the extraction of collaborative information from the raw interaction sequences,
enabling the model to deeply explore the semantic features of learners' behaviors.
\name further introduces a parameter-free pattern extraction operation that maps the meta-behavioral sequences into the global collaborative pattern matrix, carefully preserving the collaborative information.
Finally, \name provides the universal usage to inject the extracted collaborative information into various KT models for improving the performance in KT.
The main contributions of this paper are summarized as follows:

\begin{itemize}

\item We propose a new meta-behavioral sequence construction to strengthen the extraction of collaborative information in the KT scenario.

\item We provide a parameter-free strategy to extract the global collaborative information and further introduce the universal injection for various downstream KT models.

\item Extensive experiments across three real-world datasets demonstrate that \name consistently improves representative KT models built on different neural network backbones.

\end{itemize}

\section{Relation Work}
\label{Sec:rw}

\subsection{Conventional Knowledge Tracing}
The core idea of conventional KT is to utilize various techniques such as classic probabilistic models~\cite{bkt, baker2008more, pavlik2009performance}, psychometric models~\cite{irt, rasch1993probabilistic, delatorre2009dina} and neural networks~\cite{piech2015deep, shen2022assessing, nakagawa2019graph, tong2020structure} to predict the knowledge state of learners from their individual history interaction records.
Among diverse methods, neural network-based approaches~\cite{piech2015deep,yeung2018addressing,zhang2017dynamic} have attracted great attentions due to the powerful ability of representation learning.
Since the learners' history interaction records could be naturally regarded as sequence data, the sequential deep-learning methods have been widely adopted for KT task, such as LSTM-based methods~\cite{yeung2018addressing,zhang2017dynamic,abdelrahman2019, wang2021temporal} and Transformer-based methods~\cite{pandey2019self,ghosh2020context,choi2020towards, long2021tracing, shin2021saint, pu2020deep}.
Besides, a recent method~\cite{liu2024mamba4kt} attempts to introduce the State Space Models (SSMs)~\cite{gu2023mamba} into KT to improved computational efficiency for long sequences.
Though previous methods have shown remarkable performance, the inherent limitation of only modeling individual interactions inevitably overlooks the informative learning behavior pattern extracted from other learners' history records, thereby restricting the their potential for the KT task.

\subsection{Collaborative Knowledge Tracing}
The emerging collaborative KT, which integrates both individual behaviors and common learning patterns to comprehensively predict the knowledge states of the target learner, have been a promising way for further enhancing the performance of KT.
The key step of collaborative KT is to extract valuable learning behavior patterns from history interactions of all learners. 
Prior studies~\cite{minn2018deep,long2022improving,CoSKT} utilize the clustering-based strategies~\cite{minn2018deep} or sub-sequence match-based strategies~\cite{long2022improving,CoSKT} to extract the collaborative information from learners who share similar learning behaviors with the target one.
The corresponding results also indicate that incorporating the collaborative information can enhance the model performance for KT.

Nevertheless, existing methods rely on rigid rules of information extraction based on the raw interaction sequences, which fail to effectively explore the deep collaborative information among learners, as well as to facilitate the performance improvement of various KT methods.
Although a recent study~\cite{gao2024collaborative} utilizes the semantic graphs to extract the collaborative information, this strategy is developed for cognitive diagnosis, not suitable for KT.
To this end, we propose \name, which leverages a tailored extraction strategy to learn collaborative information and transform it into a plug-and-play module to service diverse KT models.

\section{Preliminaries}
\label{Sec:pre}
% \subsection{Problem Definition}

In the KT task, we have the set of learners $\mathcal{S}=\{s_1,...s_m\}$, the set of questions $\mathcal{Q}=\{q_1,...q_n\}$ and the set of knowledge concepts $\mathcal{C}=\{c_1,...c_l\}$ where $m$, $n$ and $l$ are the number of learners, questions and knowledge concepts, respectively.
Each question $q \in \mathcal{Q}$ is associated with one or multiple specific concepts $C_q \in \mathcal{C}$.
An interaction record of a learner at time step $t$ is denoted as a tuple $x_t = (q_t, C_{q_t}, r_t)$. Here, $q_t$ represents the answered question, $C_{q_t}$ is the corresponding knowledge concept set which may contain more than one skill, and $r_t \in \{0, 1\}$ denotes the binary response correctness, where $r_t = 1$ indicates a correct answer, otherwise an incorrect answer.
For a given the historical interaction sequence of the target learner from step $1$ to $t$, formulated as $\mathcal{X}_{1:t} = \{x_1, x_2, \dots, x_t\}$,
The goal of KT is to predict the response answer $r_{t+1}$ of the new question $q_{t+1}$ at the next time step $t+1$.

\section{Methodology}
\label{Sec:method}
% \subsection{Overall Architecture of MBP-KT}

\begin{figure*}[t]
    \centering
    % 0.95\textwidth
    \includegraphics[width=13cm]{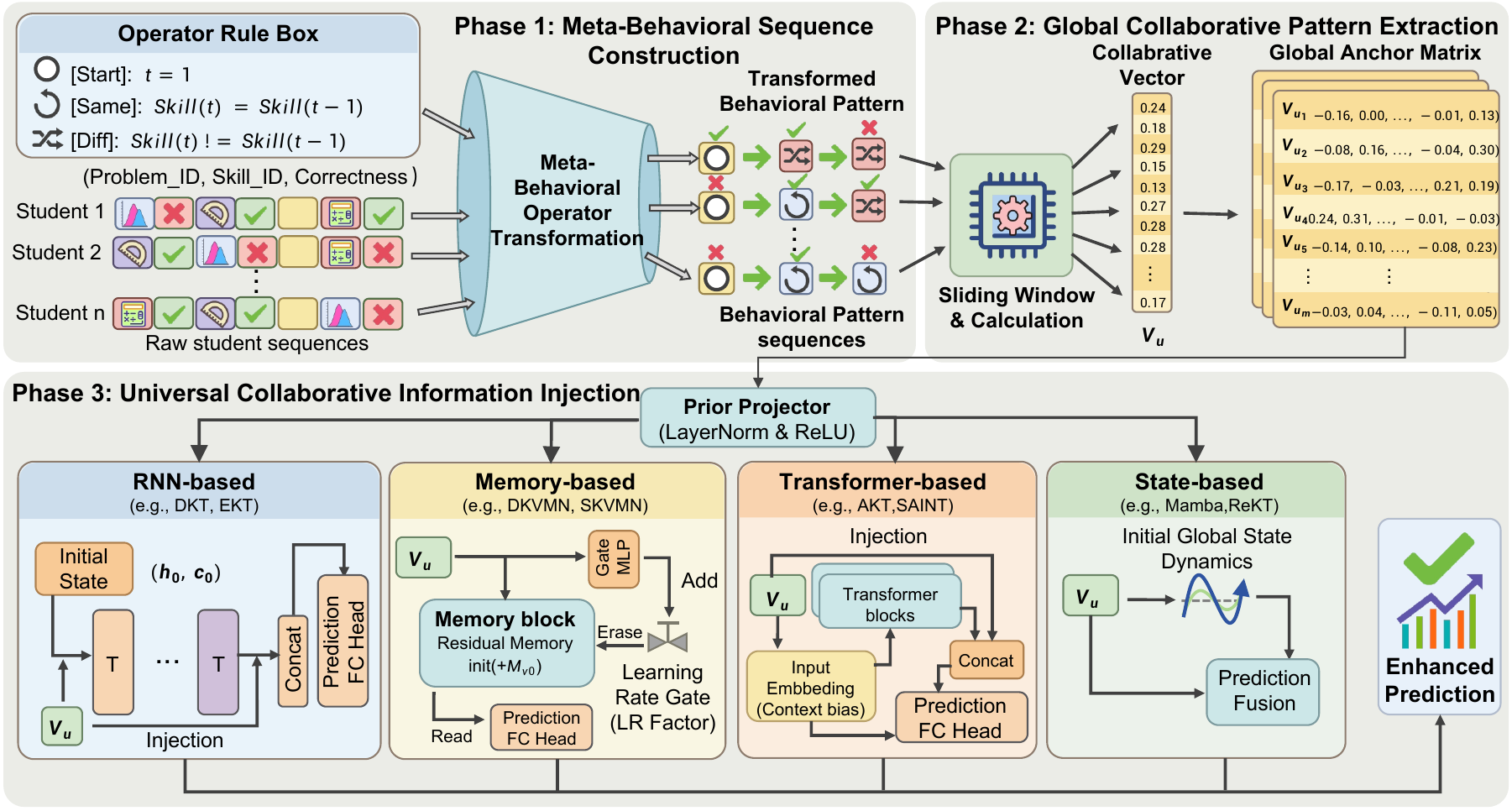}
    \caption{The overall architecture of \name. }
    \label{fig:overall_architecture}
\end{figure*}

In this section, we comprehensively introduce the proposed \name which contains three core components: meta-behavioral sequence construction, global collaborative pattern extraction and universal collaborative information injection.
The overall framework of \name is illustrated in Figure \ref{fig:overall_architecture}.

\subsection{Meta-Behavioral Sequence Construction}

To effectively extract the collaborative information from the response sequences of learners, we first develop a meta-behavioral operator transformation.
The goal of the meta-behavioral operator is to transform the raw interaction sequences into the content-agnostic meta-behavioral sequences, achieving the deep exploration of the learning behaviors of learners.

Given the raw interaction sequence $\mathcal{X}_{1:t} = \{x_1, x_2, \dots, x_T\}$ within $T$ time steps, we consider the successive learning behaviors $(x_{t-1}, x_t)$ over adjacent time steps to construct the meta-behavioral operator $\omega_t$.
Specifically, given the target time step $t$, the transformation from $(x_{t-1}, x_t)$ to $\omega_t$ is formally defined as: 
% We define a meta-behavior token $m_t$ over adjacent time steps $(x_{t-1}, x_t)$ as:
\begin{equation}
    \label{eq:relational_operator}
    \omega_t = \begin{cases} 
    \texttt{Start}, & \text{if } t=1, \\ 
    \texttt{Same}, & \text{if } C_{q_t} \cap C_{q_{t-1}} \neq \emptyset, \\ 
    \texttt{Diff}, & \text{if } C_{q_t} \cap C_{q_{t-1}} = \emptyset, 
    \end{cases}
\end{equation}

where $C_{q_t}$ denotes the knowledge concept set at time step $t$.
$\{\mathtt{Start}, \mathtt{Same}, \mathtt{Diff}\}$ is the set of operators which describes three types of states in the interaction sequence. 
The $\mathtt{Start}$ operator indicates the beginning of the interaction sequence.
The $\mathtt{Same}$ operator characterizes the learning behaviors of the target learner under similar knowledge concepts, preserving the knowledge state within a knowledge domain.
While the $\mathtt{Diff}$ operator captures the situation that the learner handles the successive questions which involves different types of knowledge concepts, exploring the cross-domain learning behaviors. 

By mapping the adjacent learning records into the different types of operators by Eq. (\ref{eq:relational_operator}), the dynamic knowledge states could be decoupled from the specific question or knowledge concepts according to the raw interaction sequence, facilitating the deep exploration of interaction records of learning behaviors.
After obtaining the meta-behavioral operators, we further combine them with the binary response correctness $r_t \in \{0, 1\}$ to construct the meta-behavioral sequence $\mathcal{Z} = \{(\omega_t, r_t)\}_{t=1}^T$.
In this way, the information of learners' meta-behavioral patterns could be carefully preserved, which serves as the key support for extracting collaborative learning behaviors.

\subsection{Global Collaborative Pattern Extraction}

Through the meta-behavioral sequence construction, the original interaction sequences are transformed into new sequences which involve three different states.
To effectively learn the collaborative information from the generated meta-behavioral sequences, we develop a global collaborative pattern extraction to reveal the collaborative patterns from interaction sequences of different learners.

The key idea of the global collaborative pattern extraction is to identify the frequently occurring behavior patterns, which could be regarded as the global collaborative information.
Specifically, we first transform all interaction sequences in the training set into the meta-behavioral sequence $\mathcal{Z}$.
Then, a sliding window strategy is adopted to analyze the combination patterns of different states:
\begin{equation}
    \label{eq:pattern}
    \mathcal{M} = \varphi (\mathcal{Z}, N, \tau),
\end{equation}
where $\varphi(\cdot)$ is the pattern extraction function, $N$ is the size of the sliding window and $\tau$ is the minimum frequency threshold for filtering out the episodic noise.
$\mathcal{M} = \{p_1, p_2, \dots, p_K\}$ denotes the various patterns of different state combinations, where $p_k = \{(\omega_i, r_i)\}_{i=1}^N$ denotes the specific state combination within the window size $N$ and $K$ is the size of $\mathcal{M}$.

The rationale of $\mathcal{M}$ is that the combinations of states $p_k$ preserve the knowledge state of the target learner.
For instance, the combination  $[(\mathtt{Same}, 0), (\mathtt{Same}, 0)]$ could indicate that the continuous struggles on the same knowledge concept.
Moreover, $\mathcal{M}$ is constructed by the interaction sequences of all learners.
Therefore, the extracted state combinations preserve the patterns of learning behavior among all learners, facilitating the exploration of common behavior patterns.

Based on $\mathcal{M}$ with extracted $K$ behavioral patterns, we further construct the global collaborative pattern matrix $\mathbf{V}\in \mathbb{R}^{m\times K}$ to represent the global collaborative information of learners.
Specifically, given the target learner $u$, the value $\mathbf{V}_{u, k}$ of the specific pattern $p_k$ in $\mathcal{M}$ is calculated as follows:
\begin{equation}
    \label{eq:v_u}
   \mathbf{V}_{u, k} = \xi (\frac{f_{u,k} - \mu_k}{\sigma_k},  \delta),
\end{equation}
where $f_{u,k} = \mathrm{Count}(p_k, Z_u)$ denotes the frequency of behavioral pattern $p_k$ in the meta-behavioral sequence of the learner $u$.
$\mu_k$ and $\sigma_k$ are the mean value and the standard deviation of the pattern $p_k$ in $\mathcal{M}$.
$\xi (\cdot)$ is the truncation function with the range $[-\delta, \delta]$ to truncate extreme outliers, improving the  robustness against behavioral anomalies.

Based on Eq. (\ref{eq:v_u}), $\mathbf{V}_{u, k}$ serves as the normalized frequency of the learner $u$ on the pattern $p_k$.
A larger value of $\mathbf{V}_{u, k}$ indicates that the corresponding behavioral pattern $p_k$ occurs more frequently in the learner’s response sequence.
Furthermore,two learners with similar learning behaviors will have similar pattern representations according to $\mathbf{V}$.
In this way, the behavioral patterns in $\mathcal{M}$ could be represented as the global collaborative pattern matrix $\mathbf{V}$, facilitating the usage of collaborative information in downstream KT models.

\subsection{Universal Collaborative Information Injection}

As a precomputed behavioral pattern feature, the global collaborative pattern matrix $\mathbf{V}$ has the potential to serve any downstream KT model.
Here, we introduce how to achieve the universal collaborative information injection in sequential model based approaches for KT via $\mathbf{V}$.
Considering the inconsistency of input feature dimensions, we first apply a projection layer to obtain the hidden representation $\mathbf{P}=\rho(\mathbf{V})$.
Then, we introduce how to enhance the performance of four mainstream sequential architectures via $\mathbf{P}$, including RNNs \cite{piech2015deep, liu2019ekt}, memory networks \cite{zhang2017dynamic, abdelrahman2019}, Transformer \cite{ghosh2020context, choi2020towards} and state-based models \cite{shen2024revisiting, gu2023mamba, liu2024mamba4kt}.

\subsubsection{Recurrent Neural Network Family}

In RNN-based architectures, such as DKT \cite{piech2015deep} and EKT \cite{liu2019ekt}, the output relies strictly on the sequential evolution of the hidden state.
However, previous methods \cite{piech2015deep, liu2019ekt} adopt the uninformative random initialization as the initial representation of the hidden state, which may degrade the model performance.
Therefore, we utilize $\mathbf{P}$ to perform meaningful state initialization:

\begin{equation}
\label{eq:rnn_init}
\mathbf{H}_0 = \mathbf{W}_h \mathbf{P}, \quad \mathbf{C}_0 = \mathbf{W}_c \mathbf{P},
\end{equation}
where $\mathbf{W}_h$ and $\mathbf{W}_c$ are projection weight matrices mapping the prior to the initial hidden state $\mathbf{H}_0$ and initial cell state $\mathbf{C}_0$, respectively. 
At the prediction stage, $\mathbf{P}$ is concatenated directly with the final hidden state:
\begin{equation}
\label{eq:rnn_output}
\mathbf{Y}_t = \sigma(\mathbf{W}_y [\mathbf{H}_t \oplus \mathbf{P}] + \mathbf{b}_y),
\end{equation}
where $\mathbf{H}_t$ is the hidden state at time step $t$, $\oplus$ denotes the concatenation operation, $\sigma(\cdot)$ represents the sigmoid activation function, and $\mathbf{W}_y, \mathbf{b}_y$ are the parameters of the prediction layer. 
This mechanism ensures that the knowledge tracking process begins with an explicit awareness of the learner's learning behavior pattern and to strengthen the role of collaborative information in the final prediction stage.

\subsubsection{Memory Network Family}

In memory network-based architectures, such as DKVMN \cite{zhang2017dynamic} and SKVMN \cite{abdelrahman2019}, knowledge states are tracked through the update of an external memory matrix.
However, standard memory architectures govern this update uniformly across all learners via shared erase ($\mathbf{e}_t$) and add ($\mathbf{a}_t$) vectors, failing to account for varying learning behavior patterns. Therefore, we first utilize $\mathbf{P}$ to perform a pattern-aware initialization of the global memory:

\begin{equation}
\label{eq:memory_state}
\mathbf{M}_{v}^{\text{init}} = \mathbf{M}_{v0} + \mathrm{Tanh}(\mathbf{W}_{\text{init}} \mathbf{P}),
\end{equation}

where the subscript $v$ explicitly denotes the Value matrix inherent to the Key-Value memory architecture, and $\mathbf{M}_{v0}$ denotes the globally shared original initial value matrix, and $\mathbf{W}_{\text{init}}$ is the corresponding learnable projection matrix. Subsequently, we introduce a pattern-aware gating mechanism:

\begin{equation}
\mathbf{M}_{v}^{new} \leftarrow \mathbf{M}_{v}^{old} \cdot (1 - w_t \mathbf{e}_t \cdot \alpha_u) + w_t \mathbf{a}_t \cdot \alpha_u, \quad \alpha_u = \sigma(\mathbf{W}_{\text{gate}} \mathbf{P}),
\end{equation}

where $\mathbf{W}_{\text{gate}}$ is the gating weight matrix. $w_t$ represents the correlation weight vector generated by the read operation at step $t$. $\mathbf{M}_{v}^{\text{old}}$ and $\mathbf{M}_{v}^{\text{new}}$ represent the value memory matrices before and after the update. The gate $\alpha_u$ acts as a personalized volatility factor. Physically, a near-zero $\alpha_u$ freezes the memory update, modeling a learner in a consolidation phase with minimal knowledge state change, while a high $\alpha_u$ enables rapid knowledge state transitions, including both fast absorption and accelerated forgetting. A residual connection is further applied at the prediction head by concatenating $p_u$ with the read-out vector before the final FC layer.

\subsubsection{Transformer Family}
For Transformer-based architectures, such as AKT \cite{ghosh2020context} and SAINT \cite{choi2020towards}, the representation learning relies on the attention matrix.
Inspired by the positional encoding techniques, we inject the collaborative prior as an additive contextual encoding directly into the input embeddings:

\begin{equation}
\label{eq:attention_state}
\tilde{\mathbf{X}} = \mathbf{X}+ \mathbf{P},
\end{equation}
where $\tilde{\mathbf{X}}$ and $\mathbf{X}$ are the augmented inputs and the original inputs, respectively.
In this way, the self-attention mechanism can carefully preserve the global collaborative information when calculating the attention matrix.
Moreover, the modified prediction layer based on Eq. (\ref{eq:rnn_output}) is also adopted for prediction stage.

\subsubsection{State Model Family}
State-based methods, such as Mamba \cite{gu2023mamba} and Mamba4KT \cite{liu2024mamba4kt}, apply the advanced sequential model SSM for KT.
However, these methods also suffer from random initialization.
To this end, we inject $\mathbf{P}$ as a persistent input-level bias, ensuring the latent state evolves under a consistent behavioral prior throughout the entire sequence:

\begin{equation}
\label{eq:state_driven_h}
\mathbf{H}_t = \mathcal{F}_{\text{state}}(\mathrm{Norm}(\mathbf{X}_t + \mathbf{P})),
\end{equation}
where $\mathcal{F}_{\text{state}}(\cdot)$ denotes the state update function of the specific backbone, and $\mathbf{X}_t$ represents the standard input feature at time step $t$. By conditioning every input step with $\mathbf{P}$, the state transition dynamics inherently preserve the global collaborative information.
Similarly, the prediction layer in Eq. (\ref{eq:rnn_output}) is applied for model prediction.

\section{Experiments}
\label{Sec:exp}

\subsection{Dataset}
In this paper, we utilize three widely-used educational datasets with different scales and  sparsity levels: ASSISTments2009\cite{feng2009addressing}, EdNet-KT1\cite{choi2020ednet}, and XES3G5M\cite{liu2023xes3g5m}.   
we adopt a single-point prediction protocol for performance evaluation in the KT task. Specifically, for each learner's trajectory, the second-to-last interaction is utilized as the validation set, and the final interaction is exclusively reserved for testing, with all preceding steps used for training.
The detailed statistics of datasets please refer to Appendix \ref{app:dataset}.

\subsection{Baseline}
We adopt representative KT models from four categories mentioned in Section 4.3.
1) RNN-based method: DKT \cite{piech2015deep}.
2) Memory network-based method: DKVMN \cite{zhang2017dynamic}.
3) Transformer-based methods: AKT \cite{ghosh2020context}, SAINT \cite{choi2020towards}, and UKT \cite{Cheng2025}.
4) State-based method: ReKT \cite{shen2024revisiting}.
Here, we also introduce the pure SSM \cite{gu2023mamba} as a baseline due to its superiority in dealing with sequential data.
The information about baselines and the implementation details about the experimental settings are reported in Appendix \ref{app:baselines} and Appendix \ref{app:implementation}, respectively.

\begin{table}[t]
\centering
\caption{Overall performance comparison of base models and their MBP-KT-enhanced counterparts across three datasets. Bold indicates the better result within each baseline-enhancement pair. $\uparrow$ indicates higher is better; $\downarrow$ indicates lower is better.}
\label{tab:main_results}
% 使用 resizebox 强制缩放到单栏文本总宽度
\resizebox{\textwidth}{!}{
\setlength{\tabcolsep}{4pt} 
\begin{tabular}{ll|cccc|cccc|cccc}
\toprule
\multirow{2}{*}{\textbf{Family}} & \multirow{2}{*}{\textbf{Models}} & \multicolumn{4}{c|}{\textbf{ASSISTments2009}} & \multicolumn{4}{c|}{\textbf{EdNet-KT1}} & \multicolumn{4}{c}{\textbf{XSE3G5M}} \\
\cmidrule(lr){3-6} \cmidrule(lr){7-10} \cmidrule(lr){11-14}
 & & AUC $\uparrow$ & ACC $\uparrow$ & RMSE $\downarrow$ & F1 $\uparrow$ & AUC $\uparrow$ & ACC $\uparrow$ & RMSE $\downarrow$ & F1 $\uparrow$ & AUC $\uparrow$ & ACC $\uparrow$ & RMSE $\downarrow$ & F1 $\uparrow$ \\
\midrule
\multirow{3}{*}{RNN} 
 & DKT & 0.7347 & 0.6853 & 0.4554 & 0.7459 & 0.7515 & 0.6806 & 0.4514 & 0.7135 & 0.8083 & 0.8254 & 0.3548 & 0.8970 \\
 & \textbf{+ MBP-KT} & \textbf{0.7618} & \textbf{0.7082} & \textbf{0.4438} & \textbf{0.7678} & \textbf{0.7768} & \textbf{0.7056} & \textbf{0.4413} & \textbf{0.7436} & \textbf{0.8152} & \textbf{0.8280} & \textbf{0.3526} & \textbf{0.8982} \\
 & \textit{$\Delta$} & \textit{\small+2.7\%} & \textit{\small+2.3\%} & \textit{\small-1.2\%} & \textit{\small+2.2\%} & \textit{\small+2.5\%} & \textit{\small+2.5\%} & \textit{\small-1.0\%} & \textit{\small+3.0\%} & \textit{\small+0.7\%} & \textit{\small+0.3\%} & \textit{\small-0.2\%} & \textit{\small+0.1\%} \\
\midrule
\multirow{3}{*}{Memory} 
 & DKVMN & 0.7567 & 0.7009 & 0.4429 & 0.7649 & 0.7521 & 0.6859 & 0.4507 & 0.7168 & 0.8003 & 0.8217 & 0.3590 & 0.8934 \\
 & \textbf{+ MBP-KT} & \textbf{0.7631} & \textbf{0.7101} & \textbf{0.4394} & \textbf{0.7753} & \textbf{0.7744} & \textbf{0.7063} & \textbf{0.4410} & \textbf{0.7379} & \textbf{0.8114} & \textbf{0.8265} & \textbf{0.3544} & \textbf{0.8969} \\
 & \textit{$\Delta$} & \textit{\small+0.6\%} & \textit{\small+0.9\%} & \textit{\small-0.3\%} & \textit{\small+1.0\%} & \textit{\small+2.2\%} & \textit{\small+2.0\%} & \textit{\small-1.0\%} & \textit{\small+2.1\%} & \textit{\small+1.1\%} & \textit{\small+0.5\%} & \textit{\small-0.5\%} & \textit{\small+0.3\%} \\
\midrule
\multirow{9}{*}{Transformer} 
 & AKT & 0.7209 & 0.6811 & 0.4538 & 0.7667 & 0.6998 & 0.6499 & 0.4701 & 0.7065 & 0.7493 & 0.8107 & 0.3733 & \textbf{0.8911} \\
 & \textbf{+ MBP-KT} & \textbf{0.7461} & \textbf{0.6965} & \textbf{0.4447} & \textbf{0.7705} & \textbf{0.7402} & \textbf{0.6777} & \textbf{0.4545} & \textbf{0.7345} & \textbf{0.7596} & \textbf{0.8107} & \textbf{0.3709} & 0.8903 \\
 & \textit{$\Delta$} & \textit{\small+2.5\%} & \textit{\small+1.5\%} & \textit{\small-0.9\%} & \textit{\small+0.4\%} & \textit{\small+4.0\%} & \textit{\small+2.8\%} & \textit{\small-1.6\%} & \textit{\small+2.8\%} & \textit{\small+1.0\%} & \textit{\small+0.0\%} & \textit{\small-0.2\%} & \textit{\small-0.1\%} \\
\cmidrule{2-14}
 & SAINT & 0.7724 & 0.7207 & 0.4336 & 0.7807 & 0.7564 & 0.6912 & 0.4486 & 0.7166 & 0.8095 & 0.8279 & 0.3541 & \textbf{0.8989} \\
 & \textbf{+ MBP-KT} & \textbf{0.8050} & \textbf{0.7403} & \textbf{0.4175} & \textbf{0.7961} & \textbf{0.7809} & \textbf{0.7069} & \textbf{0.4380} & \textbf{0.7360} & \textbf{0.8177} & \textbf{0.8283} & \textbf{0.3515} & 0.8980 \\
 & \textit{$\Delta$} & \textit{\small+3.3\%} & \textit{\small+2.0\%} & \textit{\small-1.6\%} & \textit{\small+1.5\%} & \textit{\small+2.5\%} & \textit{\small+1.6\%} & \textit{\small-1.1\%} & \textit{\small+1.9\%} & \textit{\small+0.8\%} & \textit{\small+0.0\%} & \textit{\small-0.3\%} & \textit{\small-0.1\%} \\
\cmidrule{2-14}
 & UKT & 0.7673 & 0.7147 & 0.4357 & 0.7780 & 0.7645 & 0.6938 & 0.4458 & 0.7235 & 0.8128 & 0.8289 & 0.3525 & 0.8984 \\
 & \textbf{+ MBP-KT} & \textbf{0.8175} & \textbf{0.7626} & \textbf{0.4095} & \textbf{0.8095} & \textbf{0.7951} & \textbf{0.7202} & \textbf{0.4328} & \textbf{0.7544} & \textbf{0.8188} & \textbf{0.8300} & \textbf{0.3507} & \textbf{0.8995} \\
 & \textit{$\Delta$} & \textit{\small+5.0\%} & \textit{\small+4.8\%} & \textit{\small-2.6\%} & \textit{\small+3.2\%} & \textit{\small+3.1\%} & \textit{\small+2.6\%} & \textit{\small-1.3\%} & \textit{\small+3.1\%} & \textit{\small+0.6\%} & \textit{\small+0.1\%} & \textit{\small-0.2\%} & \textit{\small+0.1\%} \\
\midrule
\multirow{6}{*}{State} 
 & ReKT & 0.7710 & 0.7147 & 0.4349 & 0.7697 & 0.7579 & 0.6922 & 0.4478 & 0.7143 & 0.8027 & 0.8241 & 0.3573 & 0.8970 \\
 & \textbf{+ MBP-KT} & \textbf{0.7916} & \textbf{0.7341} & \textbf{0.4254} & \textbf{0.7855} & \textbf{0.7837} & \textbf{0.7074} & \textbf{0.4355} & \textbf{0.7329} & \textbf{0.8155} & \textbf{0.8256} & \textbf{0.3542} & \textbf{0.8971} \\
 & \textit{$\Delta$} & \textit{\small+2.1\%} & \textit{\small+1.9\%} & \textit{\small-0.9\%} & \textit{\small+1.6\%} & \textit{\small+2.6\%} & \textit{\small+1.5\%} & \textit{\small-1.2\%} & \textit{\small+1.9\%} & \textit{\small+1.3\%} & \textit{\small+0.1\%} & \textit{\small-0.3\%} & \textit{\small+0.0\%} \\
\cmidrule{2-14}
 & Mamba & 0.7292 & 0.6930 & 0.4494 & 0.7647 & 0.7054 & 0.6556 & 0.4667 & 0.7037 & 0.7664 & 0.8138 & 0.3680 & 0.8926 \\
 & \textbf{+ MBP-KT} & \textbf{0.7789} & \textbf{0.7255} & \textbf{0.4309} & \textbf{0.7874} & \textbf{0.7733} & \textbf{0.7062} & \textbf{0.4410} & \textbf{0.7461} & \textbf{0.7759} & \textbf{0.8168} & \textbf{0.3654} & \textbf{0.8931} \\
 & \textit{$\Delta$} & \textit{\small+5.0\%} & \textit{\small+3.2\%} & \textit{\small-1.8\%} & \textit{\small+2.3\%} & \textit{\small+6.8\%} & \textit{\small+5.1\%} & \textit{\small-2.6\%} & \textit{\small+4.2\%} & \textit{\small+0.9\%} & \textit{\small+0.3\%} & \textit{\small-0.3\%} & \textit{\small+0.0\%} \\
\bottomrule
\end{tabular}
}
\end{table}

\subsection{Performance Comparison}
To comprehensively evaluate the performance of each model in KT tasks, we adopt four widely-used evaluation metrics: the Area Under the Receiver Operating Characteristic Curve (AUC), Prediction Accuracy (ACC), Root Mean Square Error (RMSE), and the F1-Score.
The detailed information about these metrics please refers to Appendix \ref{app:evametric} due to the space limitation.
The experimental results of performance comparison are reported in \autoref{tab:main_results}.

We can observe that across all four architecture families (RNN, Memory network, Transformer, and State Model), the integration of the \name yields consistent performance improvements in almost all situations. 
These results indicate that introducing the collaborative information extracted by the meta-behavioral sequences can effectively enhance the performance of diverse backbones for KT tasks, demonstrating the universality of the proposed \name.
This phenomenon also supports the claim that integrating the global collaborative information is beneficial to the prediction of individual behaviors.
In addition, it can be seen that \name yields larger performance gains on ASSISTments2009 and EdNet-KT1 compared to XES3G5M. 
The reason is that XES3G5M contains a larger number of knowledge concepts than other datasets which may influence the construction of the meta-behavioral sequences, consequently affecting the performance improvement.

\subsection{Study on Global Collaborative Pattern}
As the core component of \name,  
the global collaborative pattern extracts the valuable collaborative information via the constructed meta-behavioral sequence.
To validate the effectiveness of this module, we design two variants for collaborative information extraction:
1) Raw-KC. This strategy directly utilizes the the global co-occurrence frequency vector of knowledge concept (KC) as the injected collaborative information.
2) Normal-KC. This strategy replaces the meta-behavioral sequence in \name with the KC sequence for extracting collaborative information.
The results of variants are reported in \autoref{tab:ablation}.

We can clearly observe that compared to the above two variants, our proposed strategy can consistently improve the model performance across different backbones, showcasing the priority of the proposed global collaborative pattern in KT tasks.
Moreover, the conventional KC-based strategies fail to consistently boost model performance. The reason is that KC‑based extraction strategies face severe data sparsity issues, which hinder the effective extraction of global collaborative information from interaction sequences of learners and ultimately degrade model performance.
The additional empirical results reported in Appendix \ref{app:sparse} about the sparsity analysis also support the above claim, which further proves the necessity of devising methods that transcend the original interaction sequences to capture learners' global collaborative information.

% \begin{table}[h]
% \centering
% \caption{Ablation study of different collaborative priors on the ASSISTments2009 dataset (AUC).}
% \label{tab:ablation}
% \begin{tabular}{lcccccc}
% \toprule
% \textbf{Variant} & \textbf{DKT} & \textbf{DKVMN} 
%   & \textbf{AKT} & \textbf{SAINT} 
%   & \textbf{ReKT} & \textbf{Mamba} \\
% \midrule
% Base Model        & 0.7347 & 0.7567 & 0.7209 
%   & 0.7724 & 0.7710 & 0.7292 \\
% + Raw Skill       & 0.7395 & 0.7534 & 0.7226 
%   & 0.7651 & 0.7705 & 0.7189 \\
% + Compressed Skill& 0.7402 & 0.7601 & 0.7229 
%   & 0.7664 & 0.7752 & 0.7145 \\
% \textbf{+ Motif (Ours)} & \textbf{0.7618} & \textbf{0.7631} 
%   & \textbf{0.7461} & \textbf{0.8050} 
%   & \textbf{0.7916} & \textbf{0.7789} \\
% \bottomrule
% \end{tabular}
% \end{table}

\begin{table}[t!]
\centering
\caption{Ablation study of collaborative priors across six representative architectures on the ASSISTments2009 dataset. Both AUC and ACC metrics are reported to comprehensively evaluate performance shifts.}
\label{tab:ablation}
\resizebox{\textwidth}{!}{
\setlength{\tabcolsep}{4.5pt}
\begin{tabular}{l|cc|cc|cc|cc|cc|cc}
\toprule
\multirow{2}{*}{\textbf{Variant}} & \multicolumn{2}{c|}{\textbf{DKT}} & \multicolumn{2}{c|}{\textbf{DKVMN}} & \multicolumn{2}{c|}{\textbf{AKT}} & \multicolumn{2}{c|}{\textbf{SAINT}} & \multicolumn{2}{c|}{\textbf{ReKT}} & \multicolumn{2}{c}{\textbf{Mamba}} \\
\cmidrule(lr){2-3} \cmidrule(lr){4-5} \cmidrule(lr){6-7} \cmidrule(lr){8-9} \cmidrule(lr){10-11} \cmidrule(lr){12-13}
 & AUC & ACC & AUC & ACC & AUC & ACC & AUC & ACC & AUC & ACC & AUC & ACC \\
\midrule
Base Model         & 0.7347 & 0.6853 & 0.7567 & 0.7009 & 0.7209 & 0.6811 & 0.7724 & 0.7207 & 0.7710 & 0.7147 & 0.7292 & 0.6930 \\
+ Raw-KC        & 0.7395 & 0.6839 & 0.7534 & 0.7016 & 0.7226 & 0.6815 & 0.7651 & 0.7196 & 0.7705 & 0.7141 & 0.7189 & 0.6881 \\
+ Normal-KC & 0.7402 & 0.6934 & 0.7601 & 0.7039 & 0.7229 & 0.6813 & 0.7664 & 0.7119 & 0.7752 & 0.7187 & 0.7145 & 0.6872 \\
+\textbf{ Ours} & \textbf{0.7618} & \textbf{0.7082} & \textbf{0.7631} & \textbf{0.7101} & \textbf{0.7461} & \textbf{0.6965} & \textbf{0.8050} & \textbf{0.7403} & \textbf{0.7916} & \textbf{0.7341} & \textbf{0.7789} & \textbf{0.7255} \\
\bottomrule
\end{tabular}
}
\end{table}

\subsection{Study on learners of Different Groups}
\begin{figure}[h]
\centering
\includegraphics[width=\textwidth]{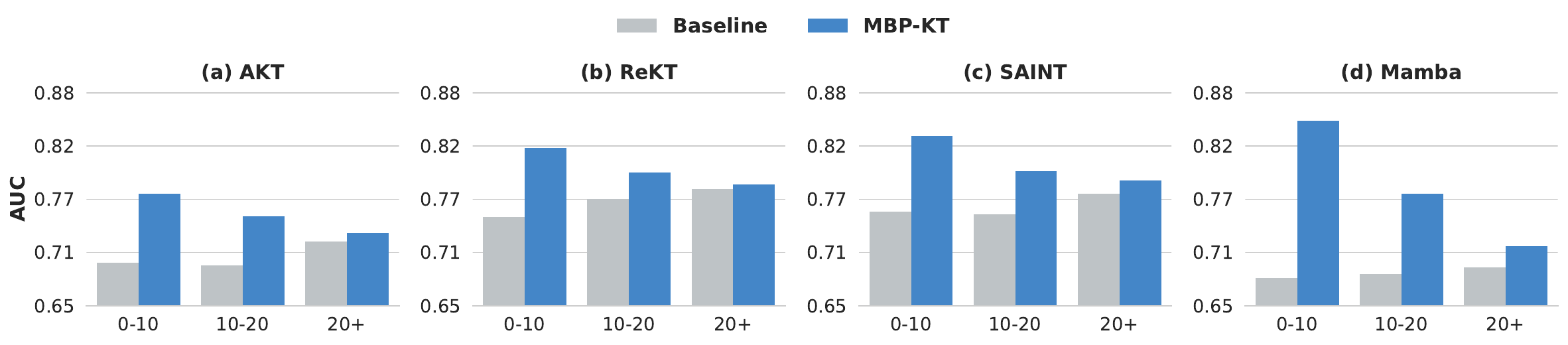}
\caption{Performance comparison across different learner groups on ASSISTments2009. 
}
\label{fig:app_coldstart}
\end{figure}

To provide deep insights of how \name enhances the performance of backbones, we evaluate the performance gain of \name on learners of different groups.
Specifically, we first divide the learners into three groups based on the number of interactions: 0-10, 10-20, and above 20.
Subsequently, we evaluate the performance gains of \name across different learner groups, and the results are shown in Figure \ref{fig:app_coldstart}.

The experimental results show that, for all models, \name yields significantly greater improvements on learner groups with fewer interaction records than on those with more records. 
The reason is that for learners with sparse interactions, global collaborative information helps effectively infer their knowledge state, whereas for those with dense interactions, their own records already offer adequate support for prediction, making the additional gain from \name relatively limited.
Moreover, to provide the comprehensive understanding of the proposed \name, we conduct additional experiments in Appendix \ref{app:inject} to analyze the influence of different injection strategies on the performance of learners with different varying interaction density.

\subsection{Study on Key Hyper-parameters}
There are two key hyper-parameters in \name: the window size of the meta-behavioral patterns ($N$) and the length of the meta-behavioral sequence ($K$).
Both of them control the quality of global collaborative information extracted from the meta-behavioral sequences.
Here, we conduct the experiments to reveal the influence of them on model performance.

\begin{figure}[h]
    \centering
    \includegraphics[width=\textwidth]{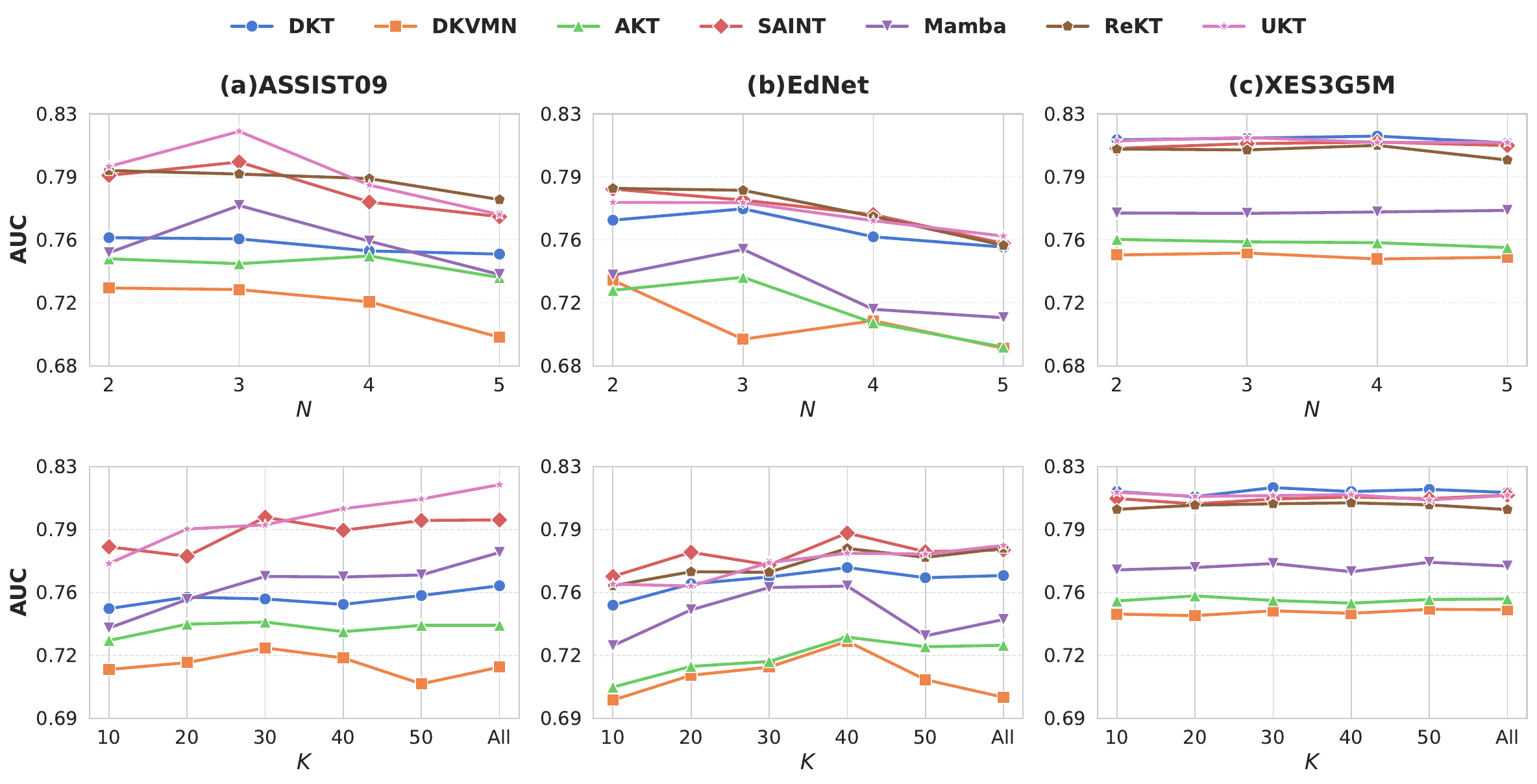} 
    \caption{Sensitivity analysis of window size $N$ and sequence length $K$.}
    \label{fig:hyper}
\end{figure}

\textbf{Study on $N$.} We evaluate the performance of \name with varying $N$ in $\{2, 3, 4, 5\}$. 
The results are shown in Figure \ref{fig:hyper}.
We can see that the optimal value of $N$ lies between 2 and 4; values that are too large or too small both lead to performance degradation. This is because when $N$ is too large or too small, the resulting meta-behavior patterns struggle to effectively capture global collaborative information, thereby reducing the benefit of incorporating such information into the model.

\textbf{Study on $K$.}
We evaluate the performance of \name with varying $K$ in $\{10,20,30,40,50,\textit{all}\}$, where $\textit{all}$ denotes we adopt all meta-behavioral patterns for collaborative information extraction. 
The results are shown in Figure \ref{fig:hyper}.
Overall, choosing a relatively large $K$ yields better performance. This is because a sufficient number of meta-behavioral patterns are required to effectively capture global collaborative information. 
However, the optimal value still depends on the characteristics of the dataset. For example, the performance variation on XES3G5M is much smaller than that on the other two datasets, which may be related to the number of knowledge concepts. Furthermore, the overall performance variation is within 2\%, indicating that model performance is not sensitive to $K$.

\section{Conclusion}
\label{Sec:con}
In this paper, we propose \name, a general framework that develops a novel global collaborative extraction module to enhance the performance of various base models in KT tasks. 
Specifically, \name introduces a new strategy that transforms the raw interaction sequences into the meta-behavioral sequence, which helps to learn the collaborative pattern from all historical response records.
Extensive results on diverse datasets show the effectiveness of \name for KT.

% \section*{Acknowledgments}
% This work is supported by . 

%\bibliographystyle{reference}/
\bibliography{reference}

%%%%%%%%%%%%%%%%%%%%%%%%%%%%%%%%%%%%%%%%%%%%%%%%%%%%%%%%%%%%
\newpage
\appendix

\section{Experimental Setup Details}

\subsection{Dataset Statistics}\label{app:dataset}

%\subsubsection{Statistics of Datasets}

To evaluate the framework across different data scales and sparsity levels, we utilize three real-world educational datasets: ASSISTments2009, EdNet-KT1, and XES3G5M. ASSISTments2009 is a classic, relatively small-scale benchmark widely used in KT literature. EdNet-KT1 is a large-scale, highly heterogeneous dataset collected from the Santa multi-platform AI tutoring service. XES3G5M is another massive real-world dataset characterized by a vast volume of user interactions and extreme sequence sparsity. 
These datasets span different educational domains and exhibit varying degrees of interaction density.
The detailed statistical information for all three datasets is summarized in Table \ref{tab:dataset_stats}. Notably, the learner-question interaction matrix sparsity , defined as $1 - \frac{\text{\# Interactions}}{\text{\# Learners} \times \text{\# Questions}}$, across all datasets remains exceptionally high, ranging from 96.26\% and peaking at 99.47\% on ASSISTments2009.

\begin{table}[h]
\centering
\caption{Descriptive statistics of the three real-world educational datasets evaluated in our experiments.}
\label{tab:dataset_stats}
\setlength{\tabcolsep}{4pt}
\begin{tabular}{lcccccc}
\toprule
\textbf{Dataset} & $\#$\textbf{Learners} & \textbf{$\#$Questions} & $\#$\textbf{KCs} & $\#$\textbf{Interactions} & \textbf{Avg. Len.} & \textbf{Sparsity} \\
\midrule
ASSISTments2009 & 5,766 & 13,441 & 150 & 412,332 & 71.51 & 99.47\% \\
EdNet-KT1       & 6,609 & 12,100 & 188 & 2,148,355 & 325.07 & 97.31\% \\
XSE3G5M         & 18,066 & 7,645 & 1,307 & 5,168,974 & 286.12 & 96.26\% \\
\bottomrule
\end{tabular}
\end{table}

To standardize the temporal modeling and accelerate experimental iterations, we restrict the maximum sequence length to $L=30$. Empirical observations indicate that a length of 30 is adequate to validate the relative performance gains of plug-and-play enhancement modules, yielding results comparable to longer sequence settings without incurring substantial computational overhead. To maximally utilize all available training data, interaction sequences in the training set exceeding this threshold are partitioned into multiple non-overlapping subsequences.

Furthermore, we adopt a single-point prediction evaluation protocol. During the training phase, models predict every time step (excluding the first) to accelerate gradient convergence. However, in the validation and testing phases, we evaluate only the final step of each sequence. This separation prevents future information leakage and ensures that the evaluation reflects the model's predictive capability in ongoing tracking scenarios.

\subsection{Baselines}\label{app:baselines}
We briefly describe the seven base models used in our experiments:

\begin{itemize}
\item \textbf{DKT (Deep Knowledge Tracing)\cite{piech2015deep}:} DKT utilizes Recurrent Neural Networks, specifically Long Short-Term Memory (LSTM) networks, to model the temporal dynamics of learner learning trajectories and continuously update hidden cognitive states.

\item \textbf{DKVMN (Dynamic Key-Value Memory Networks)\cite{zhang2017dynamic}:} This memory-augmented architecture utilizes a dual-memory structure. It stores underlying knowledge concepts in a static key matrix while dynamically updating the learner's evolving mastery levels in a corresponding value matrix through read and write operations.

\item \textbf{AKT (Attentive Knowledge Tracing)\cite{ghosh2020context}:} AKT introduces a context-aware, monotonic attention mechanism designed to capture long-range dependencies while explicitly modeling the temporal decay of historical interactions based on question and skill context.

\item \textbf{SAINT (Separated Self-Attentional Neural Knowledge Tracing)\cite{choi2020towards}:} SAINT leverages a pure Transformer-based architecture with a separated encoder-decoder structure. It independently encodes the sequence of exercises and decodes the sequence of learner responses through stacked self-attention layers.

\item \textbf{UKT (Uncertainty-aware Knowledge Tracing)\cite{Cheng2025}:} UKT models the inherent uncertainty of learner interactions by representing cognitive states as stochastic distribution embeddings, and captures the transitions between these state distributions using a Wasserstein self-attention mechanism.

\item \textbf{ReKT (Revisiting Knowledge Tracing)\cite{shen2024revisiting}:} ReKT revisits the KT task by eliminating redundant architectural components and introducing a simplified state transition mechanism for explicit cognitive state modeling, achieving competitive performance with reduced complexity.

\item \textbf{Mamba (State Space Model for KT)\cite{gu2023mamba}:} We adapt the recent selective State Space Model (Mamba) for the KT task. It efficiently captures long-range dependencies with linear temporal complexity.
\end{itemize}

\subsection{Implementation Details}\label{app:implementation}
\textbf{Experimental Environment:} All experiments were conducted on a single NVIDIA RTX 3090 GPU with PyTorch 2.3.1 and Python 3.12.

\textbf{Data Preprocessing:} To handle varying lengths of learner interaction histories, we partitioned long trajectories into multiple non-overlapping segments with a maximum sequence length of 30. Furthermore, extremely short sequences containing fewer than 5 interactions were filtered out prior to training to mitigate the impact of structural noise.

\textbf{Training Configuration:} To ensure a strictly fair comparison, all baseline models and their \name-enhanced counterparts share an identical fixed training configuration: the Adam optimizer without weight decay, a learning rate of $1 \times 10^{-3}$, a batch size of 32, and a training duration of 5 epochs. Preliminary experiments confirmed that this configuration achieves stable convergence under our data scale and sequence length settings, with additional epochs yielding no statistically meaningful improvement. Since all models are trained under this identical protocol, any observed performance gains can be exclusively attributed to the injected MBP prior rather than differences in training conditions.

\textbf{Hyperparameter Search:} Both baseline and \name-enhanced variants were tuned within the identical search space via grid search. The key hyperparameters and their search ranges are summarized in Table \ref{tab:hyperparam_space}. All architecture-specific structural parameters were kept strictly identical between each base model and its \name counterpart.

\begin{table}[h]
\centering
\caption{Hyperparameter search space utilized for model tuning.}
\label{tab:hyperparam_space}
\setlength{\tabcolsep}{12pt}
\begin{tabular}{ll}
\toprule
\textbf{Hyperparameter} & \textbf{Search Range} \\
\midrule
Embedding dimension $d_{\text{model}}$ & $\{16, 32, 64, 128\}$ \\
RNN hidden dimension & $\{16, 32, 64, 128\}$ \\
DKVMN memory slots & $\{20, 40, 50\}$ \\
Transformer layers & $\{1, 2, 4\}$ \\
Attention heads & $\{2, 4, 8\}$ \\
Dropout ratio & $\{0.1, 0.2, 0.3\}$ \\
\bottomrule
\end{tabular}
\end{table}

\subsection{Evaluation Metrics}\label{app:evametric}
The predictive performance of all models is evaluated using four standard metrics: the Area Under the Receiver Operating Characteristic Curve (AUC), Prediction Accuracy (ACC), Root Mean Square Error (RMSE), and the F1-Score. AUC serves as the primary metric due to its robustness against imbalanced label distributions, while the remaining metrics provide supplementary perspectives on classification balance and model calibration.

\textbf{Area Under the ROC Curve (AUC):} AUC measures the probability that a randomly chosen positive instance (correct response) is ranked higher by the model than a randomly chosen negative instance (incorrect response). It is formally defined as:
\begin{equation}
    \text{AUC} = P(\hat{y}_{\text{pos}} > \hat{y}_{\text{neg}}),
\end{equation}

where $\hat{y}_{\text{pos}}$ and $\hat{y}_{\text{neg}}$ represent the predicted continuous probabilities for randomly sampled positive and negative interactions, respectively. A higher AUC value indicates better discriminative ability, with 0.5 representing a random guess and 1.0 implying perfect prediction.

\textbf{Accuracy (ACC):} ACC represents the proportion of correct predictions over the total number of predictions. By applying a standard threshold of 0.5 to convert the predicted continuous probabilities into binary outcomes, it is calculated as:

\begin{equation}
    \text{ACC} = \frac{TP + TN}{TP + TN + FP + FN},
\end{equation}

where $TP$, $TN$, $FP$, and $FN$ denote the number of true positives, true negatives, false positives, and false negatives, respectively.

\textbf{Root Mean Square Error (RMSE):} RMSE quantifies the difference between the continuous predicted probabilities and the actual binary ground truth labels. A lower RMSE indicates better model calibration and more precise probability outputs. It is defined as:

\begin{equation}
\text{RMSE} = \sqrt{\frac{1}{N} \sum_{i=1}^{N} (y_i - \hat{y}_i)^2},
\end{equation}

where $N$ is the total number of predictions, $y_i \in \{0, 1\}$ is the true binary label, and $\hat{y}_i \in [0, 1]$ is the predicted continuous probability.

\textbf{F1-Score:} The F1-Score is the harmonic mean of Precision and Recall, providing a more balanced evaluation than Accuracy, particularly in scenarios where the distribution of correct and incorrect responses is skewed. It is calculated as:

\begin{equation}
\text{F1} = \frac{2 \times TP}{2 \times TP + FP + FN},
\end{equation}

A higher F1-Score indicates that the model maintains a strong balance between correctly identifying true positive responses and minimizing false positives.

\section{Supplementary Experiments}

\subsection{Analysis of Sparsity on Behavior Pattern}\label{app:sparse}

\begin{figure}[ht!]
    \centering
    \includegraphics[width=\textwidth]{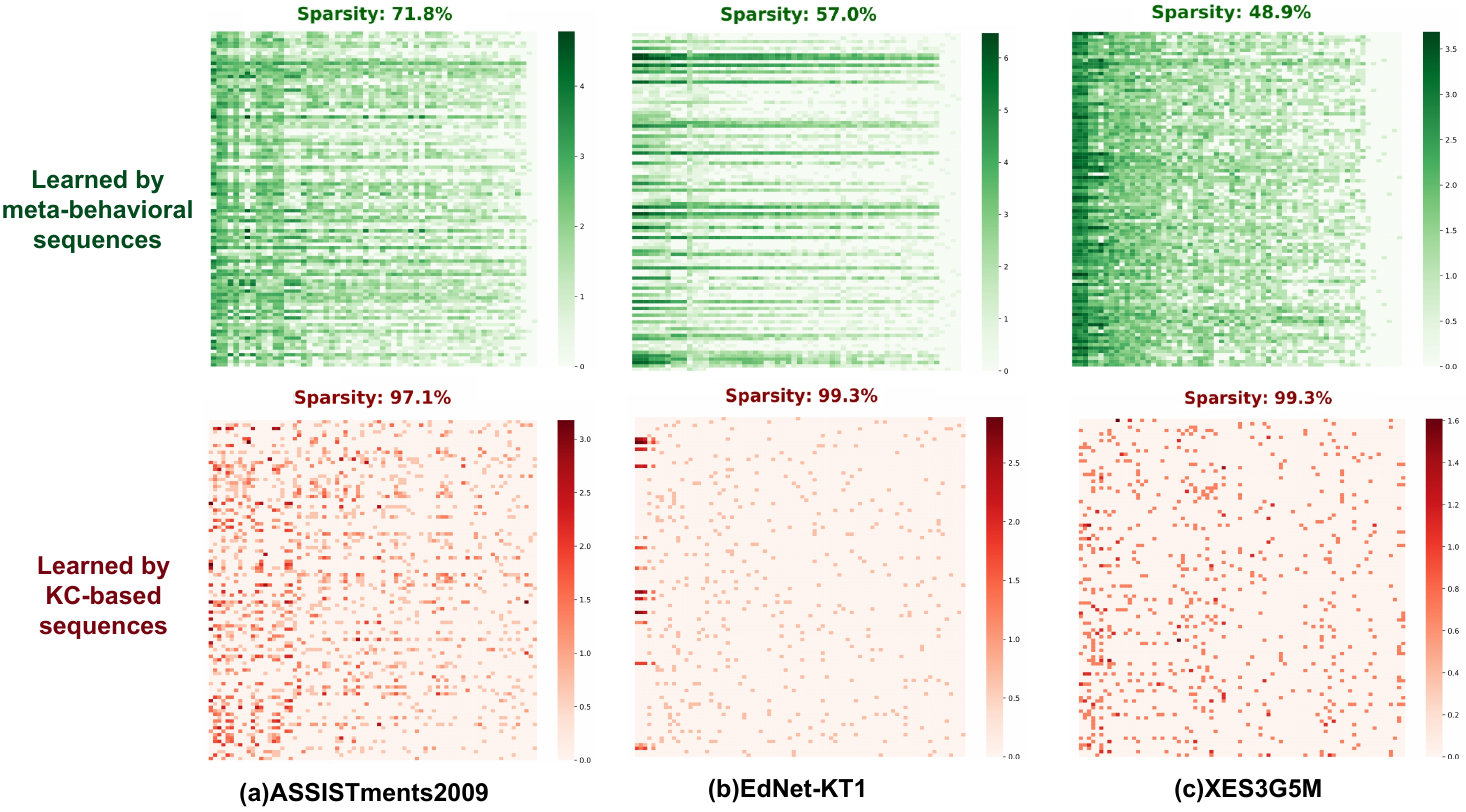} 
    \caption{ Comparison of collaborative representations' sparsity. The top row shows the representations learned by our proposed strategy, while the bottom row displays the representations extracted by conventional KC-based sequences across three datasets.}
    \label{fig:sparsity_comparison}
\end{figure}

To further validate the effectiveness of the proposed meta-behavioral sequence in extracting the global collaborative information, we conduct the experiments to analyze the collaborative representations learned by our proposed meta-behavioral sequences and traditional KC-based sequences.
Specifically, we compare the sparsity of representation vectors extracted by different strategies.
The results are illustrated in Figure \ref{fig:sparsity_comparison}.

We can clearly observe that the collaborative representations learned by the KC-based sequences exhibit extreme sparsity, reaching up to 99.3\% on EdNet-KT1 and XES3G5M. 
Such sparse distributions leave the vast majority of learner pairs with zero overlap and no viable similarity signal, failed in extracting the global collaborative information.
In contrast, our proposed strategy can significantly reduce the sparsity of collaborative representations across all datasets (e.g., from 97.1\% to 71.8\% on ASSISTments2009, and from 99.3\% to a significantly denser 48.9\% on XES3G5M).
This phenomenon demonstrates the effectiveness of proposed meta-behavioral sequence in extracting the global collaborative information.

\subsection{Study on Injection Strategies}\label{app:inject}

\name offers two strategies to inject the extracted global collaborative information. 
To validate the influence of different strategies on model performances, we compare three integration strategies: \textit{Input} (early fusion at the embedding layer), \textit{Output} (late fusion prior to the final prediction head), and \textit{Dual} (fusion at both stages).

\begin{figure}[h]
\centering
\includegraphics[width=\textwidth]{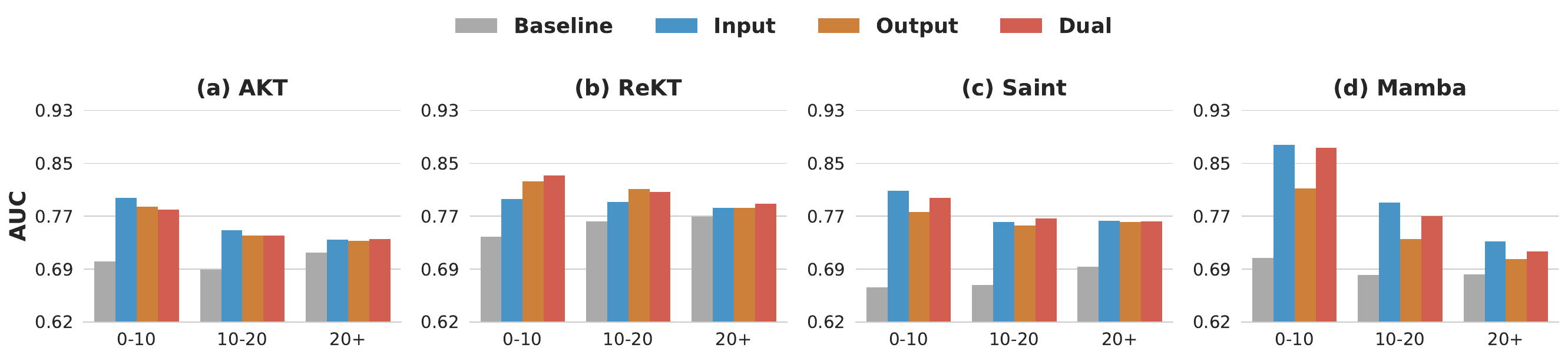}
\caption{Study on different injection strategies (Input, Output, Dual) across different sequence lengths.}
\label{fig:app_injection}
\end{figure}

Figure \ref{fig:app_injection} illustrates the performance of these three strategies compared to the baseline across four architectures. 
The primary observation is that all three injection strategies consistently outperform the baseline across varying sequence lengths, further validating the general utility of the global collaborative prior.

However, the empirical results indicate that there is no universal optimal injection position; rather, the effectiveness functions as an architecture-sensitive design choice. 
Specifically, \textit{Input} fusion demonstrates a pronounced advantage primarily in short-sequence scenarios (0-10 bins) for architectures like AKT and SAINT, where early structural guidance helps self-attention mechanisms establish initial routing. 
However, this advantage diminishes or even reverses in longer sequences (e.g., the 20+ bin for Mamba), where individual context dominates.

Conversely, the ReKT model exhibits consistent superiority when utilizing \textit{Dual} or \textit{Output} fusion across all sequence lengths. This behavior aligns with ReKT's unique state-transition mechanics: its core framework models complex memory dynamics, including temporal forgetting and knowledge state updating. A collaborative prior injected purely at the input stage may be gradually diluted during these sequential memory update cycles. In contrast, late fusion strategies (\textit{Output} or \textit{Dual}) provide a more direct, unattenuated calibration to the final predictive representation.

These architecture-dependent results demonstrate that while the global collaborative prior is universally beneficial, its optimal injection strategy should be tailored to the representational dynamics of various KT baselines.

\subsection{Deep insights in Global Collaborative Representations}
To provide a more comprehensive understanding of how \name extracts collaborative behavioral patterns,
We first visualize the learned collaborative representations across all three datasets (ASSISTments2009, EdNet-KT1, and XES3G5M).

\begin{figure}[h]
\centering
\includegraphics[width=\textwidth]{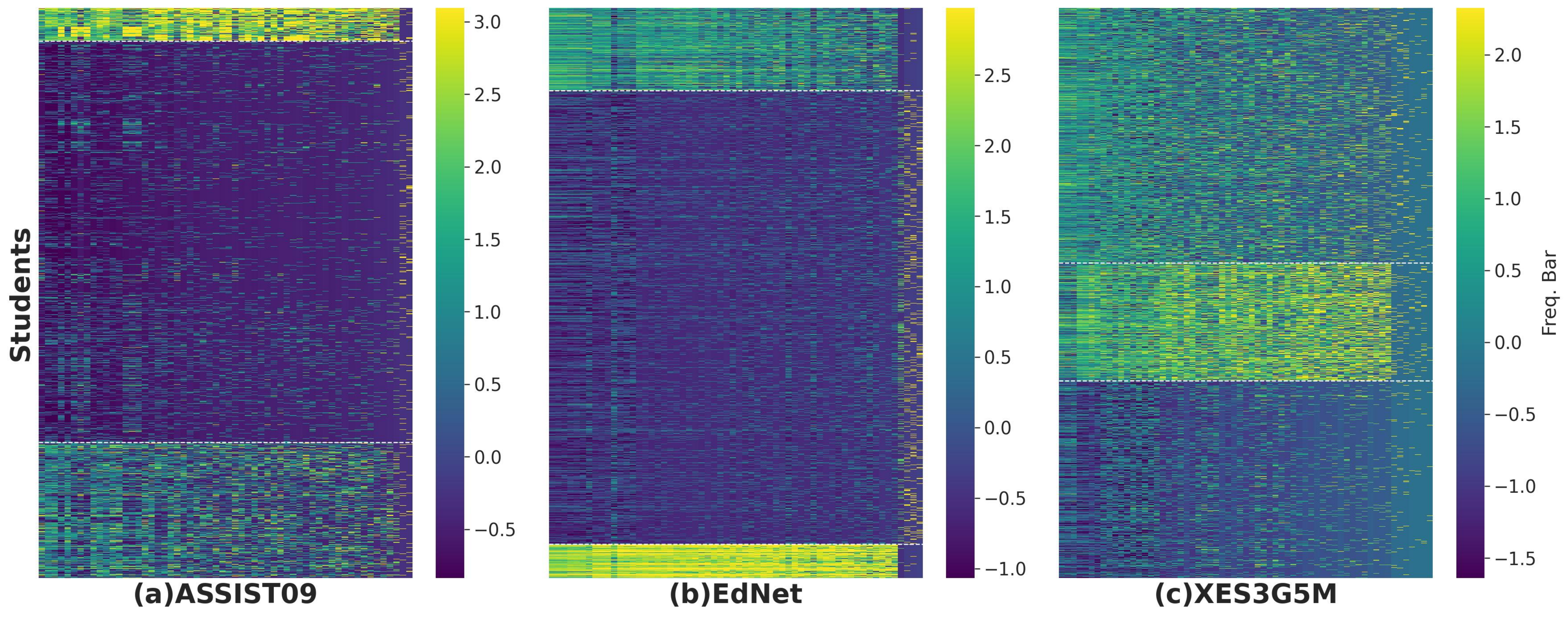}
\caption{Visualization of collaborative representations' heatmaps across datasets. (a) ASSISTments2009. (b) EdNet-KT1. (c) XES3G5M.}
\label{fig:app_heatmaps}
\end{figure}

Figure \ref{fig:app_heatmaps} visualizes the compact heatmaps of the extracted collaborative representations for learners across the three datasets. 
While the specific activation patterns vary across datasets, a consistent characteristic is the presence of several highly activated columns (corresponding to high-frequency) shared universally among learners, as well as distinct inter-group differences demarcated by the dashed lines. This provides visual evidence that \name successfully extracts the global collaborative information.

\begin{figure}[h]
\centering
\includegraphics[width=\textwidth]{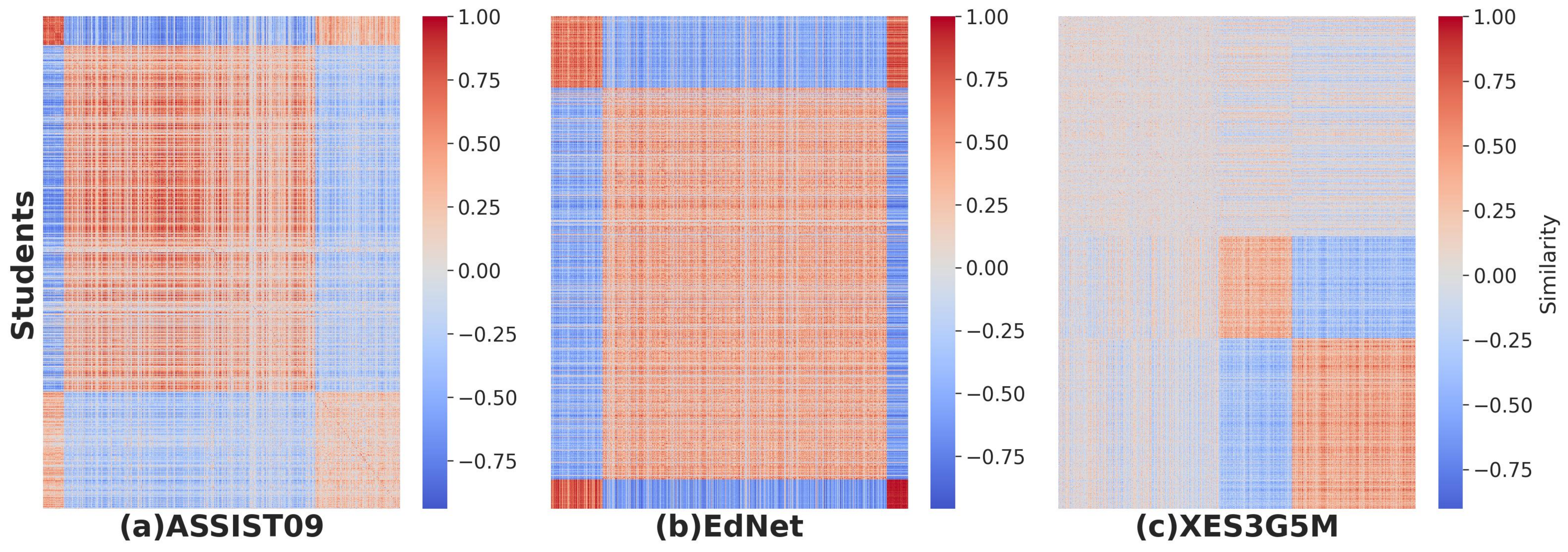}
\caption{Visualization of learner-to-learner similarity matrices derived from collaborative representations (sorted by 3 clusters). (a) ASSISTments2009. (b) EdNet-KT1. (c) XES3G5M.}
\label{fig:app_similarities}
\end{figure}

Figure \ref{fig:app_similarities} further presents the corresponding learner-to-learner similarity matrices computed directly from these collaborative representations.
We can clearly observe the cluster structures in all datasets, which provides intuitive evidence that utilizing the proposed meta-behavior sequences can effectively extract the global collaborative signals from the response records.

\subsection{Visualization of Learners' Representations}
% \textbf{Representational Clustering (t-SNE):} 

\begin{figure}[h]
\centering
\includegraphics[width=\textwidth]{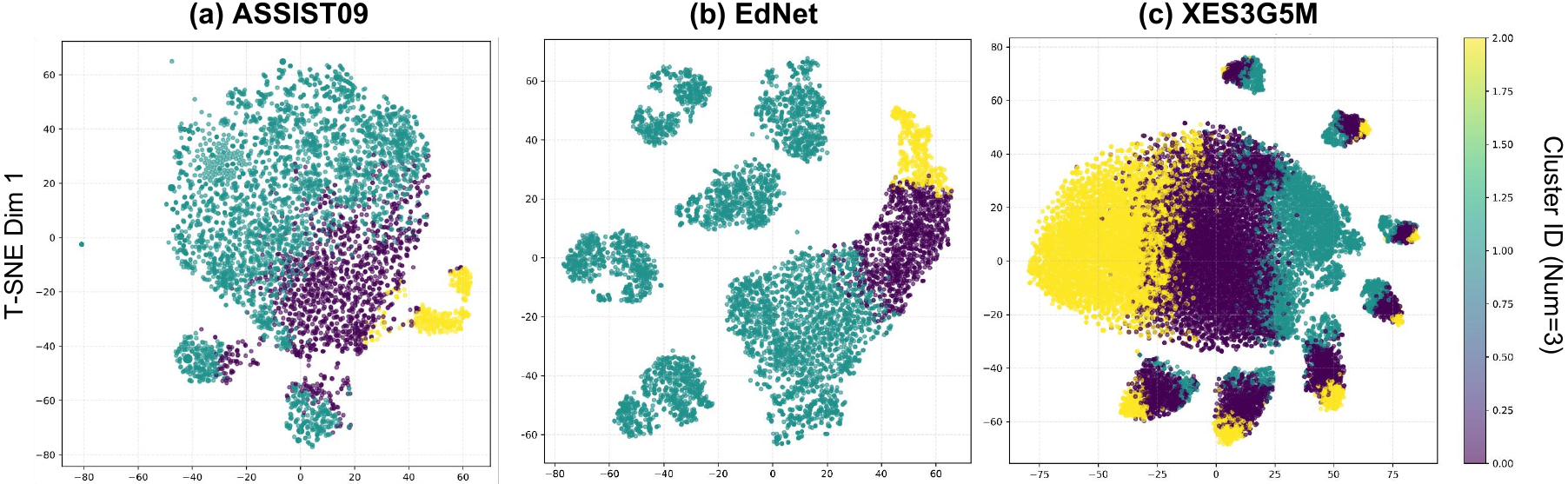}
\caption{The t-SNE visualization of learner representations derived from meta-behavioral sequences on three datasets ($K$=3).}
\label{fig:tsne}
\end{figure}

To further verify the discriminative power of the framework, we project the learned global collaborative representations of learners into a 2D space using t-SNE (Figure \ref{fig:tsne}). The visualization reveals a consistent topological pattern across datasets: a dominant, broadly dispersed primary cluster captures prevalent, mixed learning behaviors, while several smaller, tightly bound clusters emerge, corresponding to specific learner subgroups exhibiting highly consistent behavioral features. This structural composition confirms that \name effectively groups learners with similar underlying learning trajectories into shared representational regions, facilitating the downstream tasks.

\subsection{Analysis the Influence of \name on Attention Mechanism}

\begin{figure}[h]
\centering
\includegraphics[width=\textwidth]{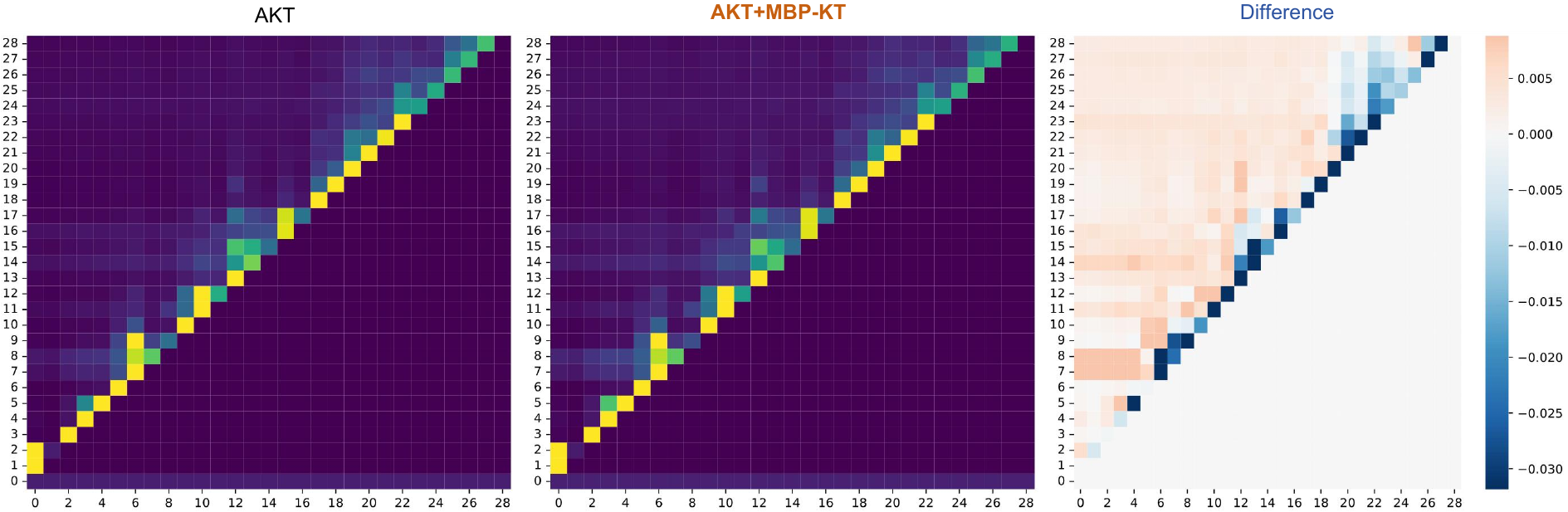}
\caption{Attention weight matrices of the baseline AKT (left), the enhanced AKT+\name (middle), and their point-wise difference (right). The difference map (Ours - Baseline) highlights the suppression of local recency bias (blue, negative) and the activation of long-range dependencies (red/orange, positive).}
\label{fig:attention}
\end{figure}

To qualitatively examine how the global collaborative information alters the model's predictive dynamics, we visualize the internal attention routing mechanisms.
Specifically, we extract the attention weight matrices from the last layer of both the baseline AKT and our enhanced variant (AKT+\name) using a sampled sequence (Figure \ref{fig:attention}). The baseline model predominantly exhibits a localized recency bias. It concentrates its attention weights densely along the diagonal (immediate past steps) while failing to retrieve relevant long-range historical context.

Upon injecting the latent pattern anchor, the attention distribution exhibits a clear structural redistribution. The difference map (Figure \ref{fig:attention}, right) visualizes this shift by plotting the enhanced variant weights minus the baseline weights. 
Negative values (dark blue bands near the diagonal) indicate a targeted suppression of redundant local attention, whereas positive values (orange/red regions in the off-diagonal areas) denote a systematic activation of distant historical interactions. 
This redistribution provides direct visual evidence that the content-agnostic topology successfully acts as a structural anchor, guiding the model to bypass sparse ID mismatches and leverage meaningful long-range collaborative signals.

\section{Limitations and Broader Impacts}
\subsection{Limitations}\label{app:limitation}
Although our proposed \name can effectively extract global collaborative information and improve the performance of various KT models, the experimental results still show that its effectiveness is not very significant in scenarios with a large number of KCs and sparse interaction records. Therefore, future research should consider designing more efficient global collaborative information extraction strategies under data-sparse conditions.

\subsection{Broader Impacts}\label{app:broaderimpacts}
Our research focuses on tracking learners' knowledge states, and the related work may be applied in smart education platforms to enhance teaching effectiveness.

%%%%%%%%%%%%%%%%%%%%%%%%%%%%%%%%%%%%%%%%%%%%%%%%%%%%%%%%%%%%

% \newpage
% \input{checklist.tex}

\end{document}